\documentclass[11pt]{article}
\usepackage[margin=1in]{geometry}

\usepackage[hidelinks]{hyperref}
\usepackage[cmex10]{amsmath}
\usepackage{amssymb,amsfonts}
\usepackage{dblfloatfix}

\usepackage[ruled,vlined]{algorithm2e}
\usepackage{graphicx}
\graphicspath{{fig/}{Figures/PDF/}{Figures/PNG/}}

\usepackage{booktabs}
\usepackage{siunitx}
\usepackage[numbers,compress]{natbib}
\usepackage{texnames}
\usepackage{bm,bbm}
\usepackage{orcidlink}

\begin{document}

\title{\textbf{Contract-Governed Training for Earth Observation: Observed Service Agreement Graphs and Coverage--Accuracy Trade-offs}}
\author{Wenzhang Du\\Department of Computer Engineering,\\Mahanakorn University of Technology International College (MUTIC),\\Bangkok, Thailand\\Email: dqswordman@gmail.com}
\date{}
\maketitle

\begin{abstract}
Earth observation (EO) models are frequently trained under implicit sampling policies that optimize global accuracy but provide no explicit guarantees on who (which regions, classes, or mission-critical strata) is being served throughout training. This paper introduces a contract-governed training paradigm for EO in which training samples are grouped into service contracts---semantically meaningful units such as (dataset, region, rare-crop indicator)---and each contract is assigned a target service share. We instantiate this paradigm as an Observed Service Agreement Graph (OSAG), a lightweight governance layer that (i) monitors contract-level exposure (coverage) during optimization, (ii) drives empirical coverage toward target shares via contract-normalized sampling weights, and (iii) exposes explicit accuracy--governance trade-offs through two knobs: a sampling mixture coefficient $\alpha$ and a contract-regularization weight $\lambda_C$. We provide a compact theory in a toy setting: OSAG sampling concentrates empirical coverage to targets; coverage deviations upper-bound service-risk deviations; and contract design (coarse vs. fine) modulates governance cost. Experiments on AVIRIS hyperspectral scenes (Indian Pines + Salinas) and multispectral Sentinel-2 EuroSAT demonstrate that OSAG can substantially reduce priority coverage error while maintaining global accuracy and improving high-priority accuracy. A EuroSAT coarse-vs.-fine contract ablation further evidences how semantically refined contracts can reduce the accuracy cost per unit of governance improvement.
\end{abstract}

\paragraph*{Keywords}Earth observation, dataset governance, sampling, hyperspectral imaging, class imbalance, fairness, contract-level coverage.

\section{Introduction}
Modern EO pipelines increasingly operate under service-driven constraints: rare crop types must be reliably monitored, urban hot-spots require prioritized coverage, and multi-region deployments demand predictable attention allocation. However, conventional training procedures implicitly ``serve'' data according to observed frequencies (or class reweighting heuristics), offering no explicit mechanism to ensure that mission-critical strata are consistently and proportionally addressed.

This motivates a governance-oriented view of training: in addition to overall accuracy, we quantify and control contract-level service coverage---how often each semantically defined contract is exposed to the learner. While adjacent areas such as class-imbalance handling \cite{cui2019}, curriculum learning \cite{bengio2009}, prioritized sampling \cite{schaul2016}, and distributionally robust training \cite{sagawa2020} offer partial solutions, they typically optimize proxy objectives (e.g., long-tail accuracy) rather than directly enforcing service agreements over semantic units relevant to EO operations.

We propose Observed Service Agreement Graphs (OSAG): contracts define nodes (optionally with adjacency edges), and training induces observed node weights via exposure counts; OSAG partitions data into contracts, monitors coverage, and enables explicit trade-offs between predictive utility and governance objectives. Our results show that:
\begin{itemize}\setlength{\itemsep}{0pt}\setlength{\parskip}{0pt}\setlength{\parsep}{0pt}
    \item OSAG can reduce priority coverage error by large margins on both hyperspectral and multispectral datasets while preserving global accuracy.
    \item OSAG improves high-priority accuracy (accuracy on priority contracts) with minimal global accuracy loss.
    \item Contract design matters: a EuroSAT coarse-vs.-fine ablation demonstrates that semantically refined contracts can reduce the governance cost, consistent with our theory.
\end{itemize}

\textbf{Contributions:}
\begin{itemize}\setlength{\itemsep}{0pt}\setlength{\parskip}{0pt}\setlength{\parsep}{0pt}
    \item Contract-governed EO training: a practical framework to define, monitor, and control contract-level service.
    \item OSAG mechanism: contract-normalized sampling with explicit trade-off knobs $\alpha$ and $\lambda_C$.
    \item Toy theory: coverage concentration, risk bound, and a contract-design effect on governance cost.
    \item Evidence on HSI/MSI: Indian+Salinas and EuroSAT, plus coarse/fine ablation.
\end{itemize}

\section{Problem Setup: Contracts, Coverage, and Service Risk}
\subsection{Contracts and target service shares}
Let training data be partitioned into $m$ contracts $C=\{1,\dots,m\}$. Each contract $c$ corresponds to a semantically meaningful unit. In our experiments:
\begin{itemize}\setlength{\itemsep}{0pt}\setlength{\parskip}{0pt}\setlength{\parsep}{0pt}
    \item Indian+Salinas (HSI): contract $\approx$ (dataset-id, spatial grid cell, rare-class flag).
    \item EuroSAT (MSI): contract $\approx$ semantic groupings of classes, with a coarse vs. fine refinement for ablation.
\end{itemize}
Each contract $c$ has $n_c$ samples and a target service share $w_c>0$ such that $\sum_c w_c = 1$. Target shares implement governance policy (e.g., higher share for rare crops or critical regions), paralleling classical resource allocation objectives where fairness or priority can be encoded via target proportions \cite{jain1984}.

\subsection{Empirical coverage and priority coverage error}
Let $C_t$ denote the contract served at training step $t=1,\dots,T$. The empirical coverage is
\begin{equation}
    \hat q_T(c) = \frac{1}{T}\sum_{t=1}^T \mathbbm{1}\{C_t=c\}.
\end{equation}
Coverage error uses the $L_1$ deviation from target:
\begin{equation}
    E_{\text{cov}}(T) = \left\lVert \hat q_T - w \right\rVert_1 = \sum_c \left|\hat q_T(c) - w_c\right|.
\end{equation}
In EO governance, the key is not only uniformity but adherence to priority targets, so we report priority coverage error relative to a priority-weighted target distribution.

\subsection{Contract-level service risk}
Let $f_\theta$ be the model and $\ell(\cdot,\cdot)$ a standard loss (cross-entropy). Define average loss on contract $c$:
\begin{equation}
    \ell_\theta(c) = \frac{1}{n_c} \sum_{(x,y)\in D_c} \ell\big(f_\theta(x), y\big).
\end{equation}
For a contract distribution $q$, define service risk
\begin{equation}
    R_\theta(q) = \sum_c q(c)\,\ell_\theta(c).
\end{equation}
This makes governance explicit: changing the coverage distribution changes the risk being optimized and the populations being effectively served.

\section{OSAG: Governance via Contract-Normalized Sampling}
OSAG implements a target contract distribution at the sampling layer. At each step:
\begin{itemize}\setlength{\itemsep}{0pt}\setlength{\parskip}{0pt}\setlength{\parsep}{0pt}
    \item sample a contract $c$ with probability $w_c$;
    \item sample uniformly within $D_c$.
\end{itemize}
Equivalently, each sample $i$ in contract $c$ receives weight proportional to $w_c/n_c$. This yields an implementable weighted sampler conceptually related to prioritized sampling \cite{schaul2016} but with policy-defined contract targets rather than TD-error heuristics.

\subsection{Trade-off knob 1: sampling mixture $\alpha$}
To avoid overly aggressive governance, we mix baseline sampling and OSAG sampling:
\begin{itemize}\setlength{\itemsep}{0pt}\setlength{\parskip}{0pt}\setlength{\parsep}{0pt}
    \item with probability $\alpha$: sample according to OSAG weights;
    \item with probability $1-\alpha$: sample from a baseline distribution (e.g., uniform over samples or a baseline policy).
\end{itemize}
This allows a continuous path between purely utility-driven training and purely contract-governed training.

\subsection{Trade-off knob 2: contract regularization $\lambda_C$}
We also support a simple contract-aware loss modulation:
\begin{equation}
    \ell_{\text{total}} = \ell_{\text{CE}} + \lambda_C \cdot \Omega(\text{priority}, \ell_{\text{CE}}),
\end{equation}
where $\Omega$ is a lightweight regularizer that increases pressure on priority contracts, echoing group-robust or constrained optimization ideas \cite{sagawa2020,agarwal2018,cotter2019}.

\section{Toy Theory (Compact): Why Coverage Control Matters}
\textbf{Proposition 1 (Coverage concentration under OSAG sampling).} Assume $\{C_t\}$ are i.i.d. with $\Pr(C_t=c)=w_c$. Then $\hat q_T(c)\to w_c$ almost surely for each $c$, and for any $\epsilon>0$,
\begin{equation}
    \Pr\big(|\hat q_T(c)-w_c|\ge \epsilon\big) \le 2\exp(-2T\epsilon^2).
\end{equation}
By union bound, $\Pr(E_{\text{cov}}(T)\ge m\epsilon)\le 2m\exp(-2T\epsilon^2)$. Implication: OSAG is a principled mechanism to drive coverage error down as training progresses, which is exactly what PrioCovErr measures.

\textbf{Proposition 2 (Service-risk deviation is bounded by coverage error).} Assume a bounded loss (e.g., 0--1 error or clipped cross-entropy) so that $0\le \ell_\theta(c)\le B$ for all $c$; then for any two contract distributions $q,\tilde q$,
\begin{equation}
    \big|R_\theta(q)-R_\theta(\tilde q)\big| \le B \left\lVert q-\tilde q \right\rVert_1.
\end{equation}
Implication: once a final contract-loss vector is realized, coverage error directly upper-bounds how far actual service risk departs from the intended policy.

\textbf{Proposition 3 (Contract design modulates governance cost).} Introduce a contract adjacency graph $G=(C,E)$. If contract losses are $\beta$-Lipschitz on $G$ in shortest-path distance, then letting $c^\star=\arg\min_c \ell_\theta(c)$ yields
\begin{equation}
    \max_c \ell_\theta(c) \le \ell_\theta(c^\star)+\beta\,\text{diam}(G),
\end{equation}
and thus
\begin{equation}
    \big|R_\theta(q)-R_\theta(\tilde q)\big| \le \big(\ell_\theta(c^\star)+\beta\,\text{diam}(G)\big)\left\lVert q-\tilde q \right\rVert_1.
\end{equation}
Implication: governance cost depends not only on how many contracts exist but on how semantically coherent they are (captured in $\beta$) and their induced structure. A refined contract design can reduce $\beta$ by reducing within-contract heterogeneity, potentially lowering the accuracy penalty per unit of governance improvement---exactly what our coarse-vs.-fine EuroSAT study probes.

\section{Experimental Protocol}
\subsection{Datasets}
Indian Pines + Salinas (AVIRIS hyperspectral) are widely used HSI benchmarks \cite{green1998aviris,plaza2009,ghamisi2017}. Bands are aligned across scenes by selecting the shared subset (standard practice). EuroSAT-MSI (Sentinel-2 multispectral) is a common EO benchmark \cite{helber2018,drusch2012}. We use the 13-band MSI setting with a consistent vector representation.

\subsection{Representation and model (architecture-orthogonal evaluation)}
To isolate the governance effect from backbone choice (CNN/ViT/etc. \cite{zhu2017}) rather than stronger spectral--spatial CNNs \cite{li2017,zhong2018}, we use a simple consistent classifier family:
\begin{itemize}\setlength{\itemsep}{0pt}\setlength{\parskip}{0pt}\setlength{\parsep}{0pt}
    \item HSI: per-labeled pixel spectral vector (aligned to common bands across scenes).
    \item EuroSAT MSI: per-patch band-mean vector (13-D).
\end{itemize}
A lightweight MLP is trained with AdamW \cite{loshchilov2019} (Adam \cite{kingma2015} as the base formulation). This design ensures that improvements in coverage and high-priority performance can be attributed to the governance layer rather than model capacity.

\subsection{Contract definitions and priority targets}
HSI (Indian+Salinas): contracts capture dataset identity, coarse geography (grid cell), and rare-class indicator; rare or critical subgroups receive higher target shares. EuroSAT: contracts capture semantic class groupings; for Proposition 3, we implement a coarse contract set and a fine refinement (more contracts), then compare governance arrows baseline$\rightarrow$OSAG.

\subsection{Baselines and OSAG policies}
We compare:
\begin{itemize}\setlength{\itemsep}{0pt}\setlength{\parskip}{0pt}\setlength{\parsep}{0pt}
    \item Rand: uniform random sampling (implicit frequency-based service).
    \item CB: class-balanced sampling (inverse-frequency) \cite{cui2019}.
    \item OSAG: contract sampling with $\alpha=1,\lambda_C=0$.
    \item OSAG-mix: $\alpha$-mix with $\alpha=0.5$ and $\lambda_C=0$.
    \item $\lambda$-fairloss: $\alpha=1$ and $\lambda_C=1$.
\end{itemize}

\subsection{Metrics}
We report Acc$_\text{all}$ and Acc$_\text{high}$ (in \%), and PrioCovErr(\%) $=100\cdot\lVert \hat q_T-w\rVert_1$ (lower is better), averaged over three seeds.

\section{Results and Analysis}
\begin{figure}[t]
    \centering
    \includegraphics[width=\linewidth]{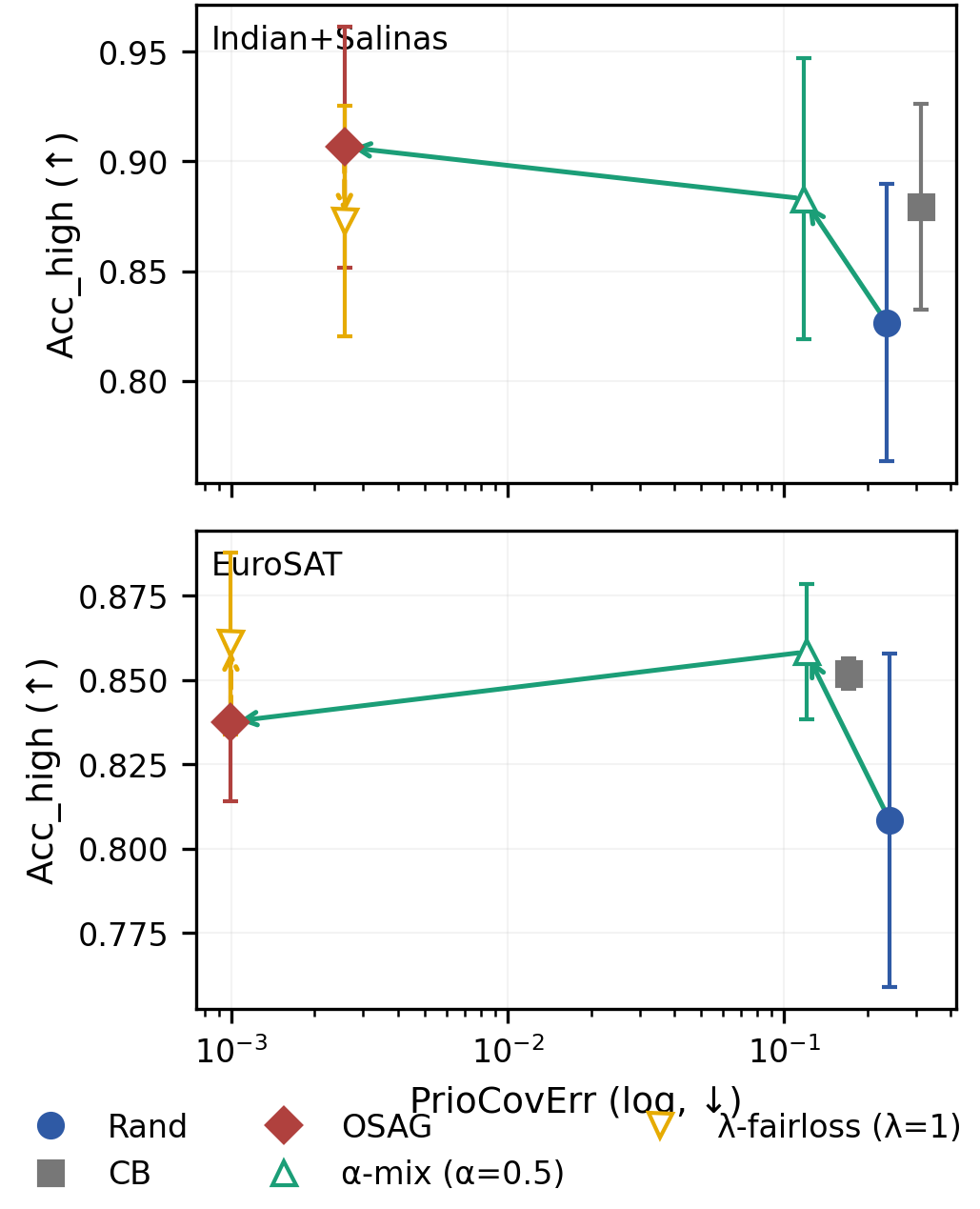}
    \caption{Governance trade-off on Indian+Salinas and EuroSAT: Acc$_{\text{high}}$ vs. PrioCovErr (log, lower is better). Error bars show std over 3 seeds.}
    \label{fig:tradeoff}
\end{figure}

\subsection{Main trade-off results (Fig.~\ref{fig:tradeoff} and Table~\ref{tab:main})}
Fig.~\ref{fig:tradeoff} shows governance trade-off curves on Indian+Salinas and EuroSAT. The $x$-axis is priority coverage error (log scale, lower is better), and the $y$-axis is high-priority accuracy (higher is better). Markers show Rand, class-balanced (CB), OSAG, OSAG-mix, and $\lambda$-fairloss points. Table~\ref{tab:main} summarizes the key operating points. OSAG improves governance (lower PrioCovErr) while maintaining global accuracy and improving high-priority accuracy.

\begin{table}[t]
\caption{Operating points (mean$\pm$std, 3 seeds). Rare: bottom 20\% frequency; priority $\in\{1,3\}$ (rare=3); targets $w_c\propto \text{priority}_c n_c$; Acc$_\text{high}$: contracts with priority=3.}
\label{tab:main}
\centering
\small
\begin{tabular}{lccc}
\toprule
Method & Acc$_{\text{all}}$ (\%) & Acc$_{\text{high}}$ (\%) & PrioCovErr (\%) \\
\midrule
\multicolumn{4}{c}{Indian+Salinas} \\
\midrule
Rand       & $91.4\pm0.3$ & $82.7\pm6.3$          & $23.49\pm0.02$ \\
CB         & $90.0\pm1.3$ & $87.9\pm4.7$          & $31.35\pm0.05$ \\
OSAG-mix   & $91.2\pm0.5$ & $\mathbf{88.3\pm6.4}$ & $\mathbf{11.76\pm0.05}$ \\
\midrule
\multicolumn{4}{c}{EuroSAT} \\
\midrule
Rand       & $84.1\pm0.2$ & $80.8\pm4.9$          & $24.08\pm0.07$ \\
CB         & $84.6\pm0.5$ & $85.2\pm0.4$          & $17.13\pm0.05$ \\
OSAG-mix   & $84.8\pm0.5$ & $\mathbf{85.8\pm2.0}$ & $\mathbf{12.01\pm0.05}$ \\
\bottomrule
\end{tabular}
\end{table}

Key observations: (i) Governance improves: PrioCovErr drops by $\approx2\times$ on both datasets, consistent with Proposition~1. (ii) High-priority performance improves: Acc$_\text{high}$ increases on both datasets. (iii) Global accuracy remains competitive: Acc$_\text{all}$ changes only marginally under governance constraints.

\begin{figure}[t]
    \centering
    \includegraphics[width=\linewidth]{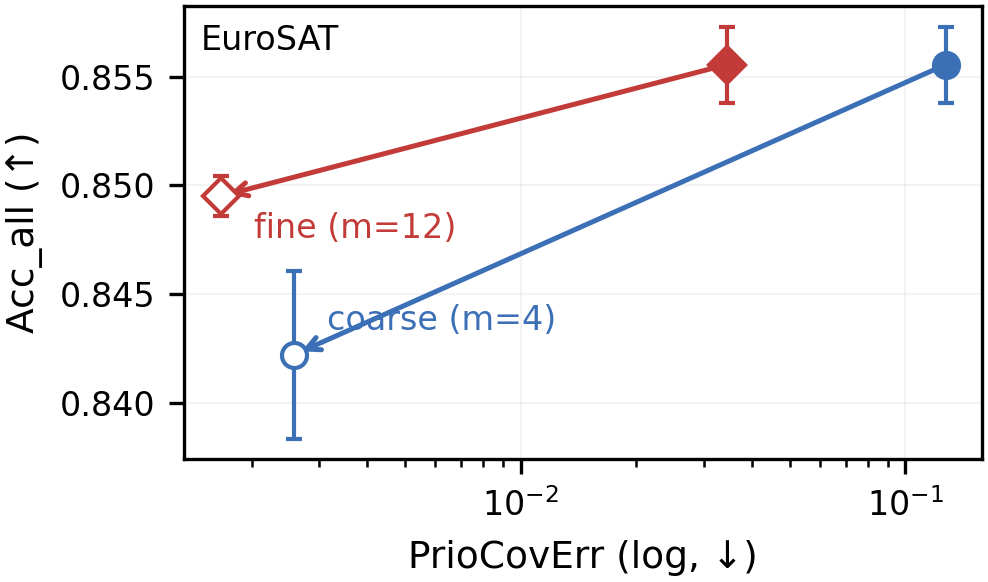}
    \caption{EuroSAT coarse vs. fine contracts: baseline $\rightarrow$ OSAG arrows in (PrioCovErr, Acc$_{\text{all}}$).}
    \label{fig:coarsefine}
\end{figure}

\subsection{Contract design matters: EuroSAT coarse vs. fine (Fig.~\ref{fig:coarsefine})}

The coarse-vs.-fine ablation supports Proposition~3: governance cost is influenced by contract structure. The fine design achieves comparable governance gain with a smaller empirical accuracy drop, consistent with Proposition~3's contract-design effect.

\section{Discussion, Limitations, and Practical Guidance}
\begin{enumerate}
    \item OSAG is governance, not a new backbone. Experiments intentionally use a simple MLP to show that governance gains are not tied to a specific architecture; OSAG can be combined with standard EO backbones (CNNs, transformers, spectral-spatial networks) that already achieve strong accuracy \cite{zhu2017}.
    \item Contract design is part of the method. As highlighted by Proposition~3 and Fig.~\ref{fig:coarsefine}, contracts should align with operational semantics (region, mission strata, rare-but-important classes). Overly arbitrary partitions can increase within-contract heterogeneity or mismatch priorities and may increase the accuracy cost of governance.
    \item Relation to imbalance and group robustness. Class-balanced methods \cite{cui2019} address label imbalance but do not enforce service targets over semantic groups. Group-robust methods optimize worst-case group risk \cite{sagawa2020} and fairness notions such as equal opportunity \cite{hardt2016}, while OSAG targets a policy-defined service distribution and monitors its compliance explicitly.
    \item Limitations. The current study covers HSI and MSI, but not SAR or multimodal city-scale benchmarks. Extending OSAG to additional modalities and larger-scale geographic partitions is an important next step, as is integrating contract-graph construction from spatial adjacency or learned similarity.
\end{enumerate}

\section{Conclusion}
We presented OSAG, a contract-governed training layer for EO that formalizes and controls who is served during training. With explicit targets over semantic contracts, OSAG reduces priority coverage error substantially, improves high-priority accuracy, and maintains competitive global accuracy on both HSI (Indian+Salinas) and MSI (EuroSAT). A EuroSAT coarse-vs.-fine ablation further shows that contract design modulates governance cost, consistent with a compact theoretical scaffold. OSAG offers a practical path toward policy-driven, accountable EO model training.

\end{document}